\documentclass[letterpaper, 10 pt, conference]{ieeeconf}  

\usepackage{stfloats}  
\newif\ifarxiv
\ifdefined\ARXIVBUILD \arxivtrue \else \arxivfalse \fi

\IEEEoverridecommandlockouts                              

\usepackage{amsmath,amssymb}
\usepackage[final]{graphicx}
\usepackage[table]{xcolor}
\usepackage{url}
\usepackage[colorlinks=true,linkcolor=blue!60!black,citecolor=blue!60!black,urlcolor=blue!60!black]{hyperref}
\graphicspath{{figures/}}

\newcommand{\NPUSHT}{30}
\newcommand{\NLIBERO}{5}

\newcounter{algorithm}
\newenvironment{algorithmbox}[1]{%
  \refstepcounter{algorithm}%
  \par\medskip
  \small
  \hrule
  \smallskip
  \noindent\textbf{Algorithm \thealgorithm: #1}
  \par\smallskip
  \hrule
  \smallskip
}{%
  \smallskip
  \hrule
  \par\medskip
}

\title{\LARGE \bf
PILOT: Guiding Long-Horizon World-Model Planning for Robot Manipulation with Visual Subgoals and Policy Priors
}

\author{Yang Liu$^{*}$, Hanchen Cui$^{*}$, and Karthik Desingh%
\thanks{$^{*}$Yang Liu and Hanchen Cui contributed equally to this work.}%
\thanks{The authors are with the Department of Computer Science and Engineering,
        University of Minnesota, Minneapolis, MN 55455, USA.
        {\tt\small \{liu03222, cui00191, kdesingh\}@umn.edu}}%
}

\begin{document}

\maketitle
\thispagestyle{empty}
\pagestyle{empty}

\begin{abstract}

Latent world-model planning has shown promising results in goal-conditioned robotic manipulation, but remains challenging for long-horizon planning due to limited temporal guidance and inefficient action search. A distant visual goal provides little guidance for preparatory actions, while uninformed action sampling makes local optimization inefficient. We propose \textbf{PILOT} (\textbf{P}olicy-\textbf{I}nitialized \textbf{L}ocal \textbf{O}ptimization with \textbf{T}emporal subgoals), which uses a video model to generate intermediate visual subgoals and a learned policy to initialize the cross-entropy method (CEM). CEM refines candidate action sequences toward each subgoal using an action-conditioned world model in a receding-horizon control loop. In our primary real-robot evaluation across four UR5e manipulation tasks, PILOT achieves 68.8\% success, compared with 55.0\% without policy initialization, 13.8\% for policy-only execution, and 0.0\% for direct final-goal planning. Simulation results show that these gains persist over longer horizons: PILOT exceeds direct execution of the same policy by 30 percentage points on long Push-T tasks and by 22 percentage points on LIBERO-10. Policy initialization also accelerates local optimization: on Push-T, CEM matches the cost reached by generic initialization after 10 iterations in only 4 iterations and reduces the final planning cost by 24.1\% under the same sampling budget. These results show that combining temporal subgoals with policy-guided action search improves both long-horizon task execution and local planning efficiency. Videos are available on the webpage: \url{https://pilot-planner.github.io/}

\end{abstract}

\section{INTRODUCTION}

World models~\cite{ha2018world,bi2026motus,guo2026ctrl,hafner2025mastering} enable robots to optimize actions at test time by predicting the outcomes of candidate behaviors. This capability is appealing for goal-conditioned manipulation, where desired outcomes vary across tasks and environments. Extending world-model planning to long horizons, however, remains challenging. Many tasks require an ordered sequence of interactions, such as approaching, grasping, transporting, and placing an object, whose intermediate stages are not explicitly specified by the final goal.

Visual world-model planners typically roll out candidate action sequences through a learned dynamics model and select them by their predicted distance to a goal observation. A distant goal image specifies \emph{what} the final scene should look like, but provides little guidance on \emph{how the task should progress}. The planner must infer both the intermediate interaction sequence and the actions needed to realize it.

\begin{figure}[t]
    \centering
    \includegraphics[width=\columnwidth]{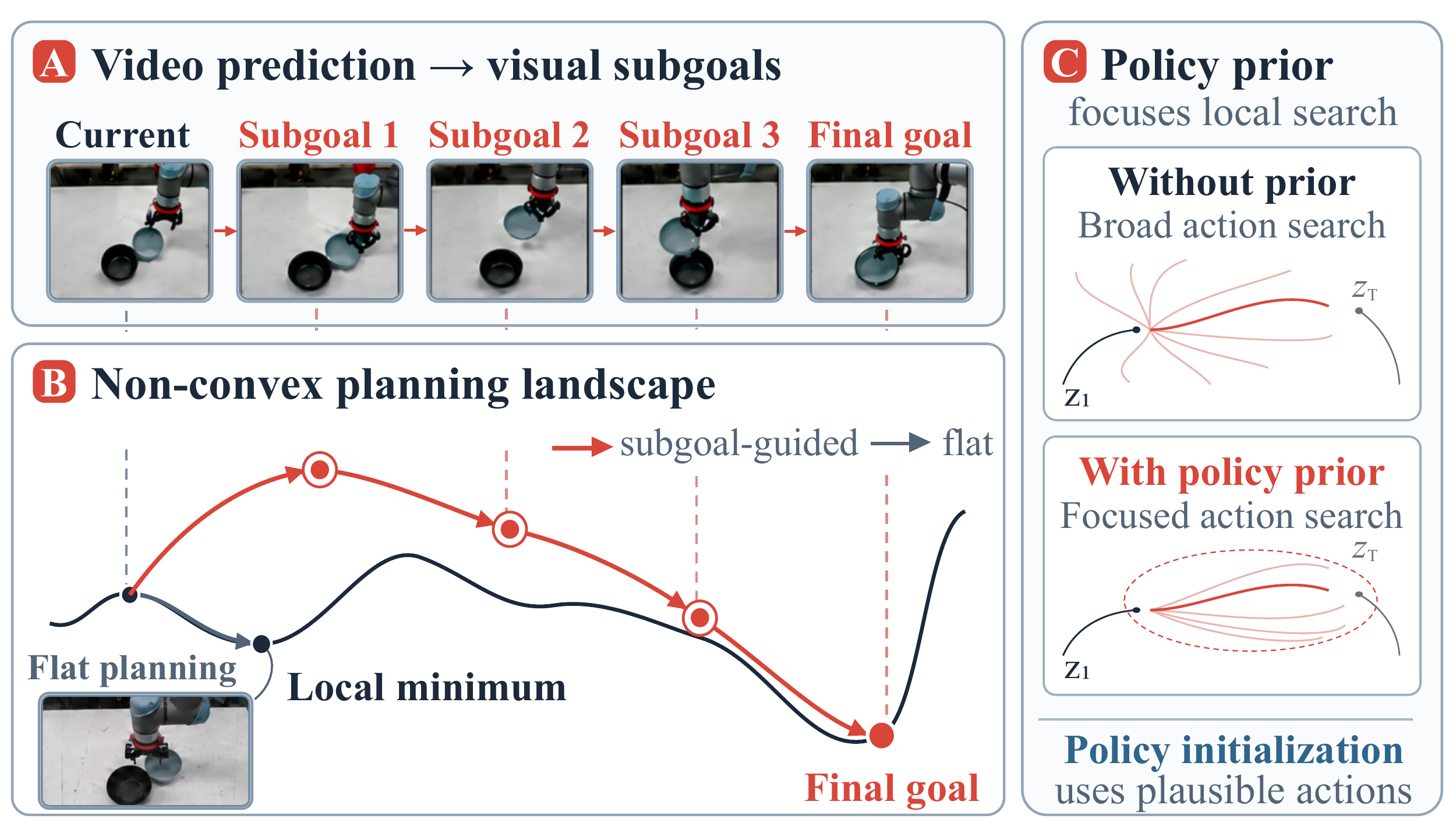}
    \caption{
    Motivation for PILOT. Visual subgoals decompose long-horizon planning into local planning stages, while a learned policy prior focuses action search on plausible trajectories.
    }
    \label{fig:motivation}
\end{figure}

The task horizon and prediction horizon play different roles. A long task can be executed through many short model-predictive control (MPC) updates, while accumulated prediction error and computational cost limit the useful prediction horizon of each update. This short horizon is what makes the final-goal objective difficult. Completing a task often requires preparatory actions, such as reaching for and grasping an object, that temporarily increase visual distance to the goal image before reducing it. A planner that looks only a few steps ahead cannot see past this temporary increase, so immediate reduction of goal distance dominates its objective. It therefore settles in a local minimum, a state that resembles the goal without completing the manipulation, as illustrated in Fig.~\ref{fig:motivation}. Reaching the global minimum requires first moving uphill in this non-convex landscape, which more effective local action search alone cannot achieve. Visual subgoals address this by placing intermediate subgoals within the available prediction horizon, so that each local problem descends toward the next stage of progress.

We use \emph{long-horizon} to describe the action planning problem rather than the task itself: task execution extends beyond a single primitive motion and beyond the prediction horizon of one local planning call. Our real-world tasks comprise multiple interaction phases within a single manipulation objective, and we study how visual subgoals guide progression across these phases.

\begin{figure*}[t]
    \centering
    \includegraphics[width=\textwidth, trim=20 30 10 1,clip]{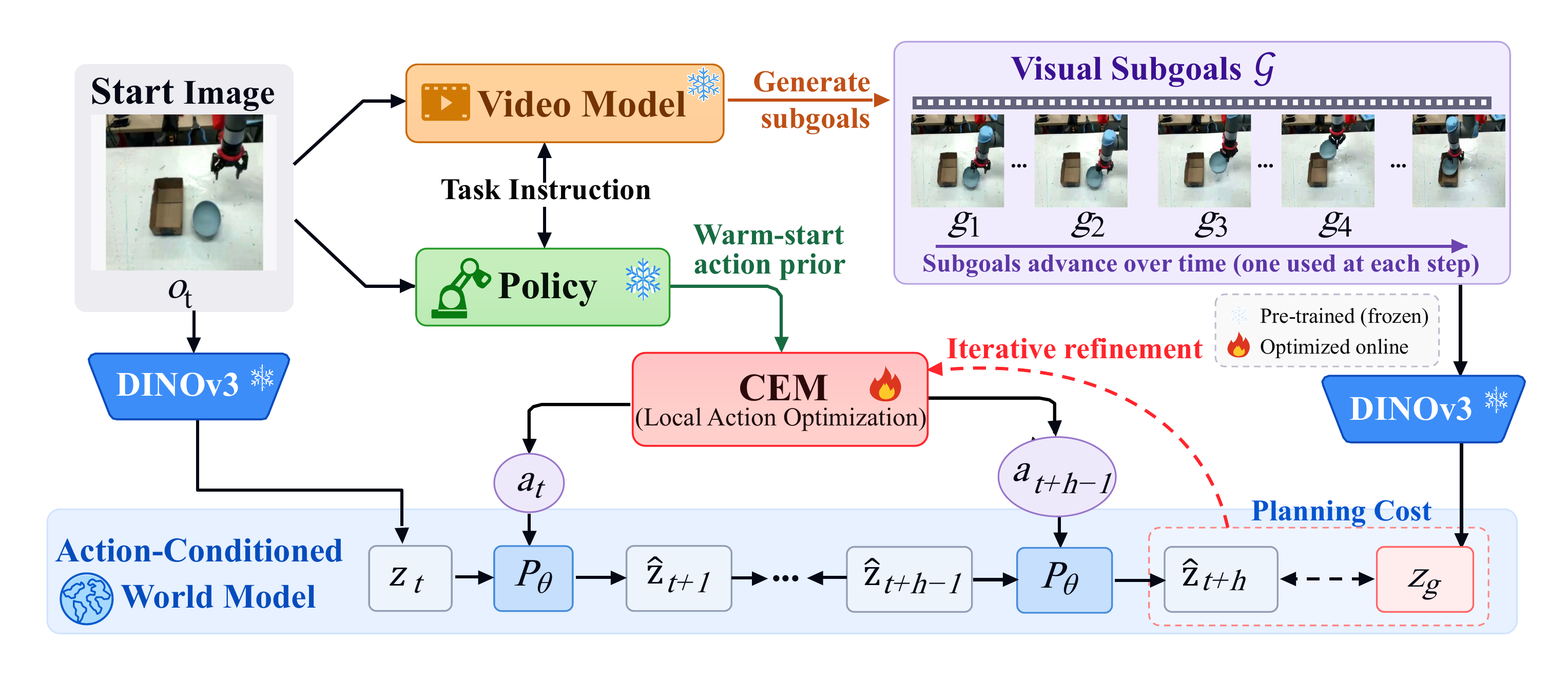}
    \caption{
    Overview of PILOT.
    A video prediction model generates visual subgoals from the current observation and task instruction.
    For each active subgoal, a learned policy initializes CEM, which refines candidate action sequences using an action-conditioned world model and a latent-space subgoal cost.
    }
    \label{fig:method_overview}
\end{figure*}

A second challenge arises from action search. Sampling-based optimizers such as the cross-entropy method (CEM) avoid differentiating through the dynamics model, but their efficiency depends on the region of action space being explored. Uninformed sampling wastes computation on implausible or irrelevant trajectories, and the search space grows rapidly with the prediction horizon.

We address these complementary difficulties with \textbf{PILOT} (\textbf{P}olicy-\textbf{I}nitialized \textbf{L}ocal \textbf{O}ptimization with \textbf{T}emporal subgoals), a hierarchical framework that separates task progression from local action optimization, as illustrated in Fig.~\ref{fig:motivation}. A video prediction model generates intermediate visual subgoals, and a learned policy initializes CEM around plausible action sequences. CEM then refines these sequences toward the active subgoal using predictions from an action-conditioned world model. The video model supplies temporal guidance, the policy focuses action search, and online world-model planning selects the actions to execute.

Our primary evaluation tests adaptation from a small set of real-robot demonstrations. We fine-tune Wan~\cite{wan2025wan} and $\pi_0$~\cite{black2024pi0} on the same 40 expert demonstrations, using Wan to propose visual subgoals and $\pi_0$ to propose actions. Across four UR5e manipulation tasks, PILOT achieves 68.8\% average success, outperforming direct final-goal planning and policy-only execution. The ablations show that generated subgoals can guide online control even when direct execution of the policy remains unreliable, and that policy initialization further improves success and local search efficiency.

Simulation complements this evaluation with controlled analyses over more initial states. Push-T varies task duration at a fixed prediction horizon; both Push-T and LIBERO-10 include final-goal planning with policy initialization and an expert-subgoal reference. These comparisons test whether policy initialization can compensate for missing temporal guidance, how prior strength affects the required planning budget, and how subgoal source affects execution.

The contributions of this work are summarized as follows:

\begin{itemize}
    \item We identify two complementary difficulties in long-horizon world-model planning: insufficient guidance for intermediate task progress and inefficient search over a large action space under a non-convex objective.

    \item We propose \textbf{PILOT}, which combines video-predicted subgoals, policy-initialized CEM, and online action refinement to execute long tasks through short-horizon world-model planning.

    \item We evaluate the contributions of visual subgoals and policy initialization to real-world task success and planning efficiency. In simulation, we further show that even weak priors improve the trade-off between success rate and planning time.
\end{itemize}

\section{RELATED WORK}

\subsection{World-Model Planning for Robotic Manipulation}

Visual world models support robotic planning by predicting future observations or latent states under candidate actions. DINO-WM~\cite{zhou2024dinowm} learns action-conditioned dynamics over frozen DINOv2~\cite{oquab2023dinov2} patch features and performs test-time optimization by matching predicted latent states to a visual goal. V-JEPA~2~\cite{assran2025vjepa2} builds on large-scale self-supervised video pretraining and trains an action-conditioned predictor on robot interaction data for image-goal manipulation. LeWorldModel~\cite{maes2026leworldmodel} instead learns a compact JEPA-style latent world model end-to-end from pixels to improve planning efficiency without a frozen foundation encoder.

These methods establish latent visual prediction as a basis for goal-conditioned control. PILOT adds a temporal hierarchy to this formulation: a video model proposes intermediate subgoals, and the action-conditioned world model solves local planning problems toward them.

\subsection{Visual Subgoals for Long-Horizon Planning}

Long-horizon manipulation~\cite{han2026robocerebra,mees2022calvin,zhang2025vlabench} is difficult to specify using only a final goal observation because the intermediate stages of task execution are often underdetermined. One approach is to provide additional intermediate observations. V-JEPA~2~\cite{assran2025vjepa2}, for example, can plan toward image subgoals, but such subgoals must be available externally and therefore require additional task-specific specification or data collection. Language provides a more convenient task-level interface, but does not directly encode the fine-grained visual changes required during manipulation.

PILOT uses a video prediction model to translate the current scene and language instruction into intermediate visual subgoals for downstream model-based control. Compared with HWM~\cite{zhang2026hierarchical}, which generates intermediate latent goals within a hierarchical world model, PILOT uses interpretable image-space subgoals and evaluates candidate actions online with a separate action-conditioned world model.

\subsection{Policy Priors for Model-Based Planning}

Sampling-based optimizers such as CEM~\cite{rubinstein1999cross} and MPPI~\cite{williams2015model} are attractive for world-model planning because they do not require differentiating through the learned dynamics. Their performance, however, depends strongly on the distribution from which candidate action sequences are sampled. In long-horizon planning, uninformed initialization can waste substantial computation on trajectories that are behaviorally implausible or unrelated to the task.

Recent work has therefore explored learned proposal distributions for model-based planning. WorldPlanner~\cite{khorrambakht2025worldplanner} uses a diffusion-based action sampler to generate structured candidate actions, while PiJEPA~\cite{chahe2026policy} warm-starts model-based optimization using a learned policy prior in visual navigation. PILOT integrates informed initialization with visual-subgoal-guided manipulation: a task-conditioned policy proposes actions, and CEM refines them toward the active subgoal using world-model predictions.

\section{METHOD}

We consider goal-conditioned robotic manipulation from visual observations. As shown in Fig.~\ref{fig:method_overview}, PILOT uses video-predicted subgoals to guide local planning and a learned policy to initialize action search. At each MPC update, CEM optimizes candidate action sequences using an action-conditioned world model. The controller executes a short prefix of the optimized sequence, observes the resulting state, and replans.

\subsection{Problem Formulation}
\label{sec:formulation}
At time $t$, the robot observes an image $o_t$, which is encoded into a latent state
\begin{equation}
z_t = \phi(o_t).
\end{equation}
An action-conditioned predictor $P_\theta$ predicts the next latent state
\begin{equation}
\hat{z}_{t+1} = P_\theta(\hat{z}_t, a_t),
\qquad
\hat{z}_t = z_t,
\label{eq:wm_transition}
\end{equation}
where $a_t \in \mathcal{A}$ denotes the robot action.

Given a visual goal $g$, we encode its latent representation as
\begin{equation}
z_g = \phi(g).
\label{eq:goal_encoding}
\end{equation}
In our implementation, the encoder $\phi$ and the action-conditioned
predictor $P_\theta$ are instantiated with
JEPA-WM~\cite{terver2025jepawm}.

Let $T_{\mathrm{task}}$ denote the task duration in control steps and
$h$ the prediction horizon of a single planning call. Model error and
computational cost constrain $h$, so completing a long task with
$h \ll T_{\mathrm{task}}$ requires repeated replanning.

A conventional visual MPC planner optimizes an $h$-step action sequence
toward the final goal using the latent-space objective
\begin{equation}
J(a_{t:t+h-1}; z_t, g)
=
\left\|\hat{z}_{t+h} - z_g\right\|_2^2,
\label{eq:flat_objective}
\end{equation}
where $\hat{z}_{t+h}$ is obtained by recursively applying
Eq.~\eqref{eq:wm_transition} along the candidate action sequence.
The corresponding planning problem is
\begin{equation}
a^\star_{t:t+h-1}
=
\arg\min_{a_{t:t+h-1}\in\mathcal{A}^{h}}
J(a_{t:t+h-1}; z_t, g),
\label{eq:flat_planning}
\end{equation}
where $\mathcal{A}^{h}$ denotes the space of action sequences of length
$h$. In practice, this optimization is approximately solved using CEM.

Here, a single final-goal objective must drive both task progression
and low-level control. PILOT separates these roles: video-predicted
subgoals specify task progression, and a policy proposal initializes
the local action search toward each subgoal. For a controlled
comparison, flat and subgoal-guided planning use the same prediction
horizon $h$, so performance differences are not due to longer
world-model rollouts.

\subsection{Temporal Visual Subgoals}
\label{sec:subgoal}

PILOT uses a video prediction model to propose a coarse visual progression of the task. The model provides subgoals for local planning; it is neither action-conditioned nor used to evaluate candidate controls.

We instantiate the video module with Wan2.1-1.3B~\cite{wan2025wan} and fine-tune it on task videos collected in the robot environment. Given the current observation $o_t$ and a language instruction $\ell$, the model predicts a future video containing $T_v$ frames:
\begin{equation}
\tilde{\mathcal{O}}
=
V(o_t,\ell)
=
\left\{
\tilde{o}_1,
\tilde{o}_2,
\dots,
\tilde{o}_{T_v}
\right\}.
\label{eq:video_trajectory}
\end{equation}

From the predicted video, we select $K$ visual subgoals
\begin{equation}
\begin{aligned}
\mathcal{G}
&=
\{g_1,g_2,\dots,g_K\},\\
g_k
&=
\tilde{o}_{\tau_k},
\end{aligned}
\label{eq:subgoal_selection}
\end{equation}
where $\tau_k \in \{1,\dots,T_v\}$ denotes the frame index of the $k$-th subgoal. The indices are uniformly spaced along the predicted video.

PILOT does not require the predicted video to correspond to an executable robot trajectory; action feasibility is assessed online by the action-conditioned world model. Generated subgoals can nevertheless be unreachable or poorly timed; the expert-subgoal comparisons in Sec.~\ref{sec:sim_setup} evaluate the effect of replacing these subgoals under the same controller.

\subsection{Policy Prior for Local Action Search}
\label{sec:prior}

Each subgoal still requires a search over robot actions. To focus CEM sampling on plausible action sequences, PILOT initializes the search with a learned policy prior $\pi_\psi$.

In our implementation, $\pi_\psi$ is obtained by fine-tuning a $\pi_0$ policy on expert demonstrations. The policy is conditioned on the current observation $o_t$ and task instruction $\ell$:
\begin{equation}
a^\pi_{t:t+h-1}
\sim
\pi_\psi(\cdot \mid o_t,\ell),
\label{eq:policy_prior}
\end{equation}
where the first $h$ predicted actions form the local action proposal.

The proposal $a^\pi_{t:t+h-1}$ initializes CEM and is not executed directly. The policy is conditioned on the task instruction but not on the visual subgoal $g_k$; the world-model objective supplies explicit alignment with the active subgoal. Our Flat WM + prior baseline (Sec.~\ref{sec:sim_setup}) plans directly toward the final goal with Eq.~\eqref{eq:flat_planning}, i.e., without intermediate visual subgoals, using the same policy initialization; it tests whether policy initialization alone can compensate for the absence of subgoals.

\subsection{Policy-Initialized Local Optimization}
\label{sec:planner}

Algorithm~\ref{alg:planner} summarizes the planning procedure. At the beginning of each episode, the video model generates a sequence of visual subgoals, which the controller processes in order.

\begin{algorithmbox}{PILOT: Policy-Initialized Local Optimization with Temporal Subgoals}
\label{alg:planner}

\textbf{Input:}
current observation $o_t$,
task instruction $\ell$,
video model $V$,
policy prior $\pi_\psi$,
world model $P_\theta$

\textbf{Hyperparameters:}
$K$, $h$, $h_{\mathrm{exec}}$,
$\Sigma_{\mathrm{init}}$, $R$, episode step limit

\begin{enumerate}
    \item
    $\tilde{\mathcal{O}}
    \leftarrow
    V(o_t,\ell)$

    \item
    $\mathcal{G}
    =
    \{g_1,\dots,g_K\}
    \leftarrow
    \textsc{SelectSubgoals}(\tilde{\mathcal{O}})$

    \item
    For $k=1,\dots,K$:
    \begin{enumerate}

        \item
        For $r=1,\dots,R$:
        \begin{enumerate}

            \item
            $z_t \leftarrow \phi(o_t)$

            \item
            $a^\pi_{t:t+h-1}
            \sim
            \pi_\psi(\cdot\mid o_t,\ell)$

            \item
            $(\mu_0,\Sigma_0)
            \leftarrow
            (a^\pi_{t:t+h-1},
            \Sigma_{\mathrm{init}})$

            \item
            $a^\star
            \leftarrow
            \mathrm{CEM}
            \left(
            J_k,
            P_\theta,
            z_t,
            \mu_0,
            \Sigma_0
            \right)$

            \item
            Execute
            $a^\star_{t:t+h_{\mathrm{exec}}-1}$
            and observe
            $o_{t+h_{\mathrm{exec}}}$

            \item
            $t
            \leftarrow
            t+h_{\mathrm{exec}}$

            \item
            Stop if the task is complete or the episode step limit is reached.

        \end{enumerate}
    \end{enumerate}

\end{enumerate}
\end{algorithmbox}


For the subgoal $g_k$, we define the local planning objective as
\begin{equation}
J_k\left(a_{t:t+h-1}; z_t\right)
=
\left\|\hat{z}_{t+h} - z_{g_k}\right\|_2^2,
\qquad
z_{g_k}=\phi(g_k),
\label{eq:subgoal_cost}
\end{equation}
where $\hat{z}_{t+h}$ is obtained by rolling out
Eq.~\eqref{eq:wm_transition} for $h$ steps from the current latent state
$z_t=\phi(o_t)$.

The local planner then solves
\begin{equation}
a^\star_{t:t+h-1}
=
\arg\min_{a_{t:t+h-1}\in\mathcal{A}^{h}}
J_k\left(a_{t:t+h-1}; z_t\right),
\label{eq:subgoal_planning}
\end{equation}
which we approximately optimize using CEM.

The learned policy proposal initializes the CEM distribution as
\begin{equation}
\mu_0
=
a^\pi_{t:t+h-1},
\qquad
\Sigma_0
=
\Sigma_{\mathrm{init}}.
\label{eq:policy_centered_cem}
\end{equation}
CEM samples candidate action sequences from the current distribution, evaluates them using world-model predictions and the objective function $J_k$, selects low-cost candidates, and refits the distribution. This refinement allows the planner to deviate from the policy proposal when other actions have lower predicted subgoal cost.

After optimization, the mean of the final CEM distribution is taken as the plan $a^\star$, and only its first $h_{\mathrm{exec}}<h$ actions are executed. The robot then receives a new observation and replans. In the real-world experiments, we use $h=2$ and $h_{\mathrm{exec}}=1$.

For each visual subgoal, we allocate $R$ planning--execution cycles before advancing to the next subgoal, with $R=2$ in the real-world experiments. This fixed schedule avoids a separate subgoal-completion model or a manually tuned latent-distance threshold and keeps the progression rule identical across tasks.

A complete traversal of $K$ subgoals executes $K\cdot R\cdot h_{\mathrm{exec}}$ control steps unless terminated early. The fixed schedule does not adapt subgoal timing when execution falls behind. Simulation settings are given in Sec.~\ref{sec:sim_setup}.

\section{EXPERIMENTS}
\label{sec:experiments}

We first evaluate task success and planning efficiency on a UR5e robot.
Push-T and LIBERO-10 then provide complementary analyses of task length,
policy initialization, and subgoal source.

\subsection{Compared Methods}
\label{sec:baselines}

We evaluate four core methods on the real robot and in simulation:

\begin{itemize}
    \item \textbf{Flat WM:}
    CEM plans directly toward the final goal image using the learned
    action-conditioned world model, without intermediate visual subgoals
    or policy-prior initialization. It uses the same short prediction
    horizon $h$ as PILOT.

    \item \textbf{Policy only:}
    The fine-tuned $\pi_0$ policy, the same network that PILOT uses as
    its prior, is executed directly, without online world-model
    evaluation or CEM refinement.

    \item \textbf{PILOT w/o prior:}
    The planner uses the same generated visual subgoals as PILOT, but
    initializes CEM from a generic distribution rather than from the
    learned policy prior.

    \item \textbf{PILOT:}
    The full method combines video-predicted visual subgoals,
    policy-prior initialization, and online action refinement with the
    action-conditioned world model.
\end{itemize}


\begin{figure}[t]
    \centering
    \includegraphics[width=\columnwidth]{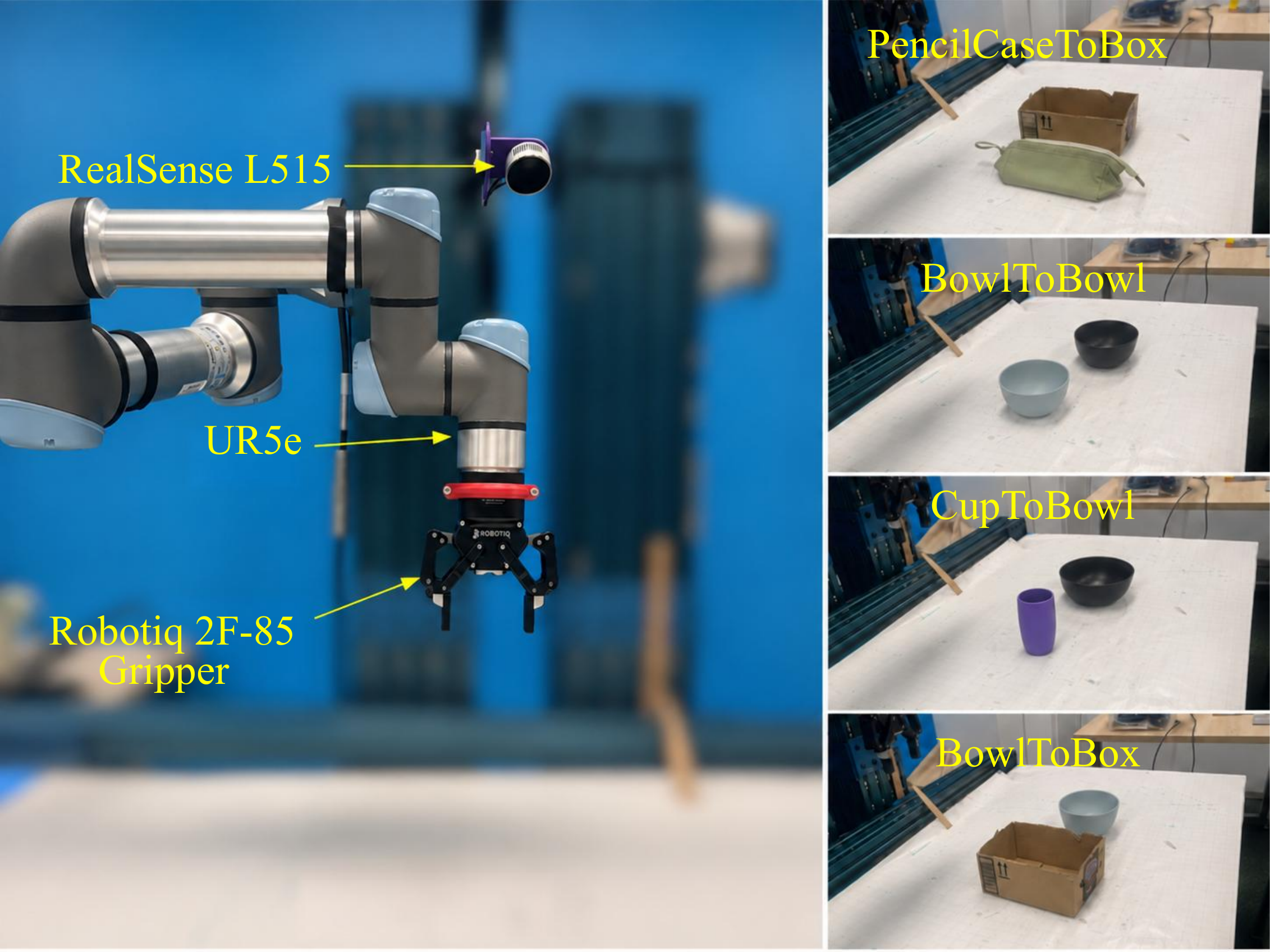}
    \caption{
    Real-world UR5e evaluation setup.
    We evaluate PILOT on four pick-and-place tasks using visual
    observations from a fixed RGB camera.
    }
    \label{fig:robosetup}
\end{figure}

\subsection{Real-World Experiments}
\label{sec:real_world}

\textbf{Setup.}
As shown in Fig.~\ref{fig:robosetup}, we use a UR5e with a fixed RGB
camera on four pick-and-place tasks: \emph{Cup} (pick up a cup and place
it into a bowl), \emph{Pencil Case} (pick up a pencil case and place it
into a box), \emph{Bowl-Box} (pick up a bowl and place it into a box),
and \emph{Bowl-Bowl} (pick up a bowl and place it into another bowl).
Each task requires approaching, grasping, transporting, and placing an
object. Actions are four-dimensional, consisting of a Cartesian
end-effector displacement $(\Delta x,\Delta y,\Delta z)$ and a gripper
command, with the end-effector orientation held fixed; each control step
executes one such action. We evaluate each method over 20 trials per task
and report per-task success counts and the overall success rate.

\textbf{Training data and implementation.}
We fine-tune Wan2.1-1.3B~\cite{wan2025wan} and
$\pi_0$~\cite{black2024pi0} on the same 40 successful demonstrations,
10 per task, using videos for Wan and observation--action trajectories
for $\pi_0$. JEPA-WM is trained on these demonstrations together with
approximately five hours of random-play interaction collected roughly
evenly across the four settings. Thus, the matched 40-demonstration
budget applies to the task-specific fine-tuning of Wan and $\pi_0$;
system training also includes world-model interaction data and
foundation-model pretraining. A trial succeeds when the robot completes
the specified manipulation within the episode step limit.

All world-model variants share the dynamics model, action bounds,
latent cost, episode step limit, and CEM budget, with $h=2$ and
$h_{\mathrm{exec}}=1$. Generated-subgoal variants use $K=6$ subgoals
and $R=2$ planning--execution cycles per subgoal. Without the policy
prior, CEM starts from a zero-mean Gaussian over the same action bounds.

\begin{figure*}[!t]
    \centering
    \includegraphics[width=\textwidth]
    {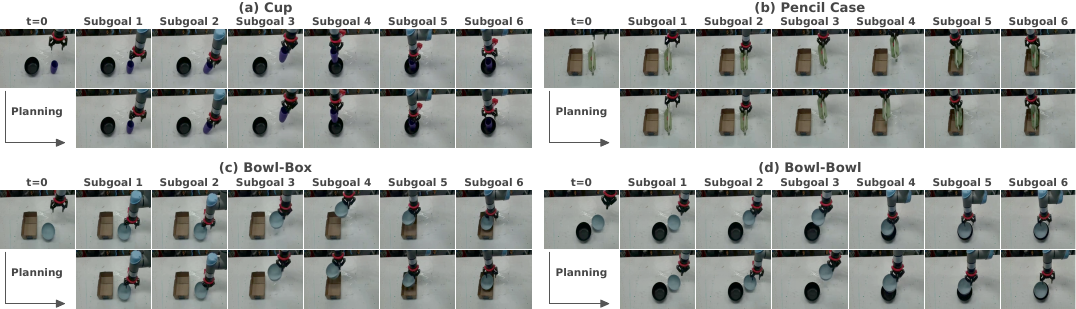}
    \caption{
    PILOT on the four real-world tasks.
    Top row of each panel: initial observation and video-predicted visual
    subgoals.
    Bottom row: robot observations after the allotted control steps for
    each subgoal.
    }
    \label{fig:real_subgoal_vs_exec}
\end{figure*}

\textbf{Task success.}
Table~\ref{tab:main_success} shows that Flat WM completes none of the
evaluated trials under the tested planning budget, while Policy only
achieves 13.8\% success. Adding visual subgoals raises world-model
planning success to 55.0\% at the same prediction horizon, supporting
the value of intermediate objectives within short rollouts. In the
human evaluation, 100\% of the generated videos on all four tasks are
judged to depict successful task completion. With Wan and $\pi_0$
fine-tuned on the same demonstrations, these subgoals guide online
control even though direct policy execution remains unreliable.

\begin{table}[t]
    \centering
    \caption{
    Real-world task success over 20 trials per task.
    Avg. pools all 80 trials; parentheses give 95\% Wilson confidence
    intervals.
    }
    \label{tab:main_success}

    \footnotesize
    \setlength{\tabcolsep}{1.7pt}

    \begin{tabular}{@{}lccccc@{}}
    \hline
    Method & Cup & \shortstack{Pencil\\Case} & Bowl-Box & Bowl-Bowl &
    Avg. (95\% CI) \\
    \hline
    Flat WM & 0/20 & 0/20 & 0/20 & 0/20 &
    0.0\% (0.0--4.6) \\
    Policy only & 4/20 & 3/20 & 2/20 & 2/20 &
    13.8\% (7.9--23.0) \\
    PILOT w/o prior & 8/20 & 5/20 & 14/20 & 17/20 &
    55.0\% (44.1--65.4) \\
    \hline
    \rowcolor{gray!12}
    PILOT & 11/20 & 13/20 & 13/20 & 18/20 &
    \textbf{68.8\% (57.9--77.8)} \\
    \hline
    \end{tabular}
\end{table}

PILOT reaches 68.8\% success, a 13.8 percentage-point gain over PILOT
w/o prior. The benefit varies by task: policy initialization adds three
successes on Cup, eight on Pencil Case, and one on Bowl-Bowl, with one
fewer on Bowl-Box. These results show an aggregate benefit from the
prior when subgoals are available; the simulation experiments also test
it with a final-goal objective.

\begin{figure}[t]
    \centering
    \begin{minipage}[t]{0.49\columnwidth}
        \centering
        \includegraphics[width=\linewidth]
        {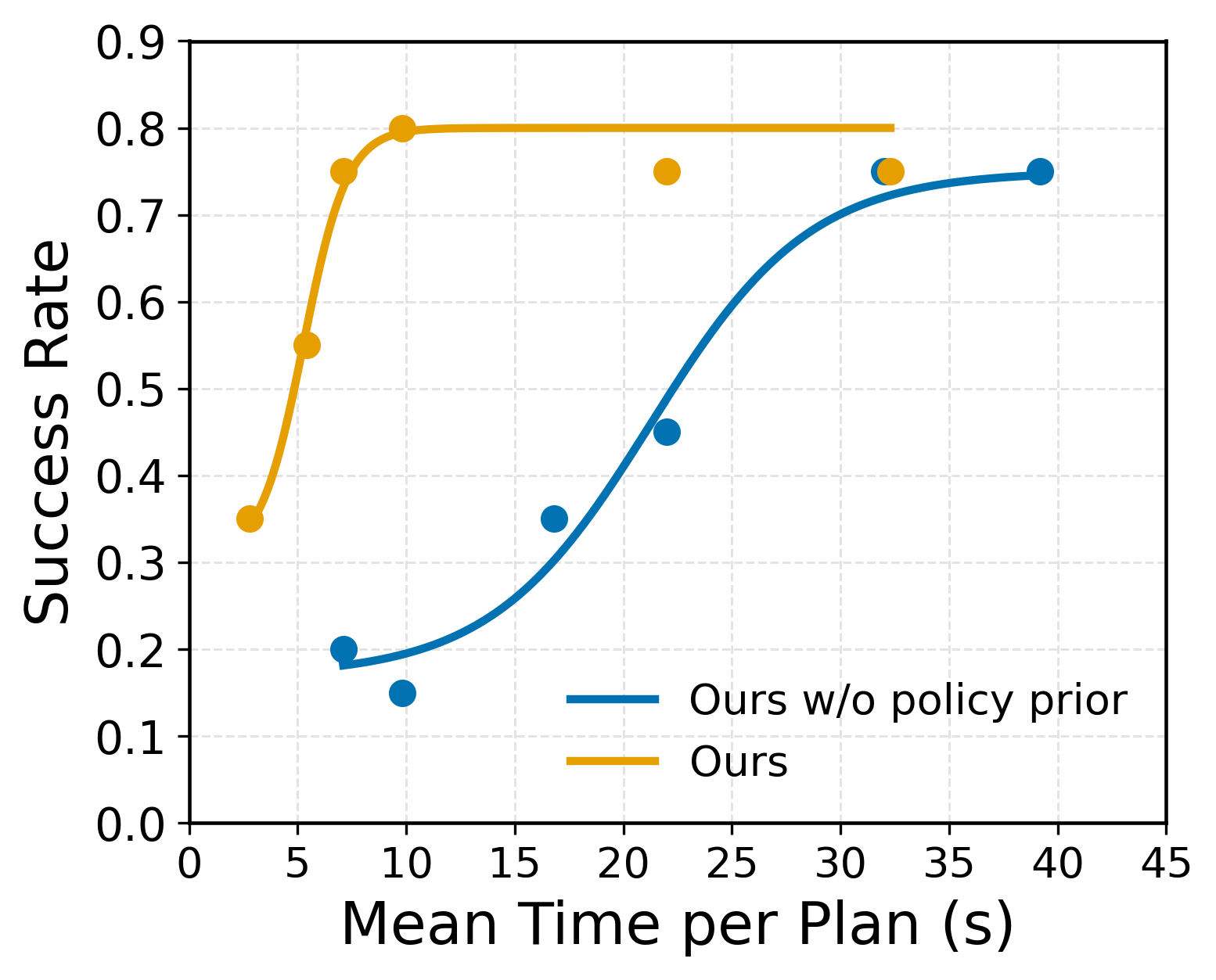}
        \small (a) Success vs.\ planning time.
    \end{minipage}
    \hfill
    \begin{minipage}[t]{0.49\columnwidth}
        \centering
        \includegraphics[width=\linewidth]
        {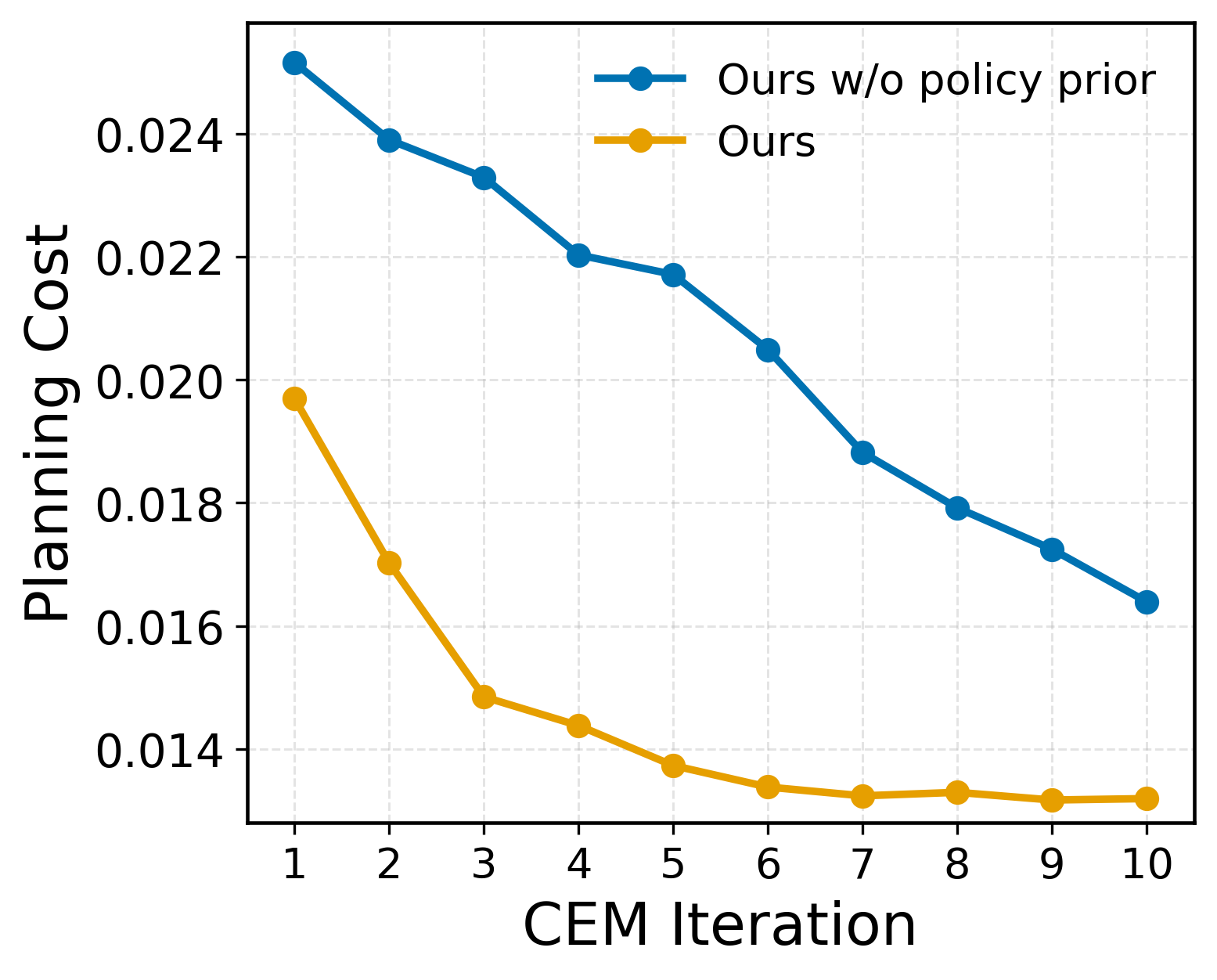}
        \small (b) Planning cost over CEM iterations.
    \end{minipage}
    \caption{
    Effect of policy-prior initialization on local planning efficiency.
    (a) Policy-guided initialization achieves higher task success under
    limited online planning budgets.
    (b) The prior provides a lower-cost initialization and maintains
    lower planning cost during CEM refinement.
    }
    \label{fig:policy_prior_efficiency}
\end{figure}

\textbf{Planning efficiency.}
We vary the online planning budget and plot success rate against mean
planning time. As shown in Fig.~\ref{fig:policy_prior_efficiency}(a),
policy initialization improves success under constrained budgets and
reaches comparable success rates with less search. The CEM curves in
Fig.~\ref{fig:policy_prior_efficiency}(b) also show lower costs at
initialization and throughout refinement, consistent with more
effective local optimization.

\textbf{Qualitative analysis.}
Fig.~\ref{fig:real_subgoal_vs_exec} shows one execution per task.
The observed states follow the subgoal sequence with a fixed, short
prediction horizon at every planning call.

\subsection{Simulation Experiments}
\label{sec:sim_setup}

\textbf{Environments.}
Push-T~\cite{chi2023diffusionpolicy} is a planar pushing task in
which a circular agent pushes a T-shaped block to a fixed target pose.
This contact-rich environment has a two-dimensional action space and
allows controlled variation of task duration.
LIBERO-10~\cite{liu2023libero} contains ten
language-conditioned, two-stage tabletop manipulation tasks performed
by a Franka arm with a seven-dimensional operational-space action
representation.

\textbf{Models and adaptation regime.}
Wan2.1-1.3B and $\pi_0$ receive environment-specific fine-tuning.
Unlike the matched demonstration budget in the real-world experiments,
simulation uses more demonstrations for Wan to adapt to rendered
scenes. On Push-T, we fine-tune $\pi_0$ on 30 expert demonstrations for
PILOT and the policy-prior baselines, and additionally train
Policy-only variants using 5, 10, 20, and 30 demonstrations. Wan is fine-tuned on 100 demonstrations. On LIBERO-10,
we use 5 demonstrations per task for $\pi_0$ and 20 for Wan.
For JEPA-WM, we use the released Push-T checkpoint and train a LIBERO
model on benchmark demonstrations and noise-perturbed replays.

Both simulation environments include two additional diagnostic
variants:
\begin{itemize}
    \item \textbf{Flat WM + prior:}
    The final-goal planner uses the same policy proposal and initial CEM
    covariance as PILOT, but retains the final goal throughout the
    episode. Together with Flat WM, PILOT w/o prior, and PILOT, this
    completes the $2\times2$ comparison of final-goal versus subgoal
    objectives and generic versus policy initialization.

    \item \textbf{PILOT (oracle):}
    The full planner replaces generated subgoals with frames from a
    held-out expert trajectory starting at the same evaluated initial
    state. The policy prior and world model are unchanged. This reference
    measures the effect of substituting expert temporal guidance; it is
    not assumed to be a strict upper bound on task success.
\end{itemize}

\textbf{Scaling with Task Horizon.}
\begin{figure}[t]
    \centering
    \includegraphics[width=\columnwidth]
    {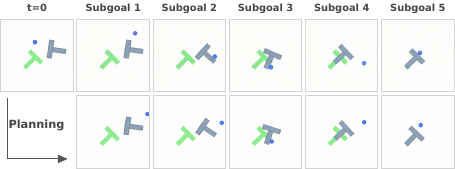}
    \caption{
    PILOT on Push-T.
    Top: initial observation and video-predicted visual subgoals.
    Bottom: simulator observations after the allotted control steps for
    each subgoal.
    }
    \label{fig:pusht_subgoal_vs_exec}
\end{figure}
We group Push-T initial states by expert episode length
$T_{\mathrm{expert}}$, measured in environment steps:
\emph{short} ($T_{\mathrm{expert}}<100$),
\emph{medium} ($100\leq T_{\mathrm{expert}}<130$), and
\emph{long} ($T_{\mathrm{expert}}\geq130$).
The observed ranges are 29--93, 110--129, and 130--238 environment
steps, respectively, with \NPUSHT{} held-out states per bucket.
One world-model step corresponds to five environment steps.
The world model and planner remain fixed, including the prediction
horizon $h$, as task duration increases.

Table~\ref{tab:pusht_main} compares all variants.
Policy only is reported for policies fine-tuned on 5 to 30
demonstrations; PILOT and Flat WM + prior use the 30-demonstration
policy as prior. Flat WM achieves 33.3\% success on short tasks and
0.0\% on both medium and long tasks. Policy initialization improves its
short-task success to 46.7\%, but medium- and long-task success remain
at 0.0\%. Under the tested planning budget, the prior therefore
improves final-goal action search without substituting for intermediate
visual subgoals on longer tasks.

Direct policy execution improves as the number of demonstrations
increases but generally degrades with task length. PILOT achieves
80.0\%, 63.3\%, and 53.3\% success on short, medium, and long tasks,
respectively, the highest rates among methods without expert subgoals.
It exceeds direct execution of the same 30-demonstration policy by
40.0, 30.0, and 30.0 percentage points and outperforms PILOT w/o prior
at every task length. These results support the complementary roles of
subgoals and policy initialization.
Fig.~\ref{fig:pusht_subgoal_vs_exec} shows a successful execution.

\begin{table}[t]
\centering
\caption{Push-T success rate by task length
(\NPUSHT{} held-out initial states per bucket).
Methods with a policy prior use the 30-demonstration policy.}
\label{tab:pusht_main}
\footnotesize
\setlength{\tabcolsep}{3pt}
\begin{tabular}{@{}lccc@{}}
\hline
Method & Short & Medium & Long \\
\hline
Flat WM              & 33.3\% & 0.0\%  & 0.0\% \\
Flat WM + prior      & 46.7\% & 0.0\%  & 0.0\% \\
Policy only (5 demos)  & 0.0\%  & 0.0\%  & 0.0\% \\
Policy only (10 demos) & 13.3\% & 3.3\%  & 6.7\% \\
Policy only (20 demos) & 26.7\% & 16.7\% & 10.0\% \\
Policy only (30 demos) & 40.0\% & 33.3\% & 23.3\% \\
PILOT w/o prior      & 43.3\% & 36.7\% & 40.0\% \\
\hline
\rowcolor{gray!12}
PILOT                & \textbf{80.0\%} &
                       \textbf{63.3\%} &
                       \textbf{53.3\%} \\
\hline
PILOT (oracle)       & 90.0\% & 93.3\% & 83.3\% \\
\hline
\end{tabular}
\end{table}

\begin{figure}[t]
    \centering
    \begin{minipage}[t]{0.49\columnwidth}
        \centering
        \includegraphics[width=\linewidth]{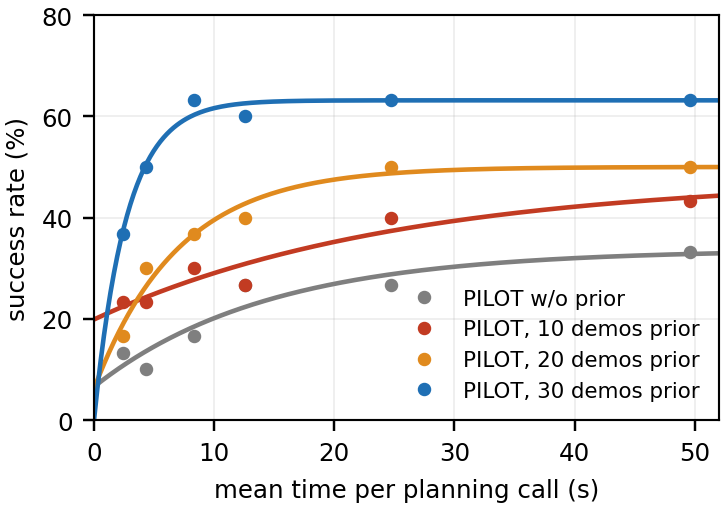}
        \small (a) Success vs.\ planning time.
    \end{minipage}
    \hfill
    \begin{minipage}[t]{0.49\columnwidth}
        \centering
        \includegraphics[width=\linewidth]{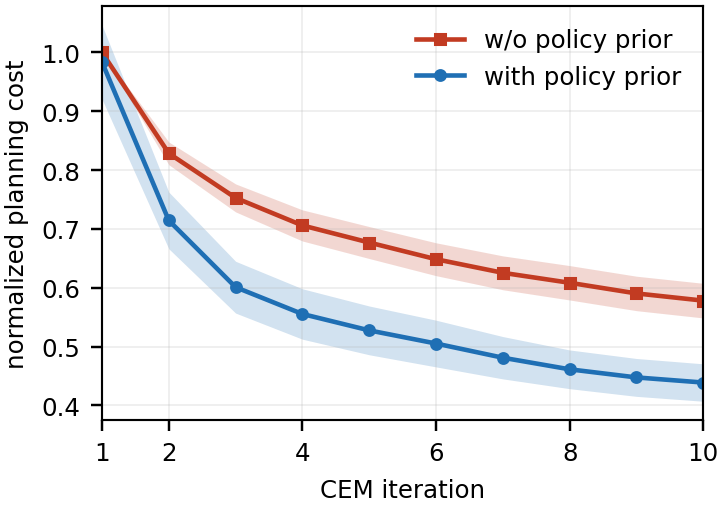}
        \small (b) Planning cost over CEM iterations.
    \end{minipage}
    \caption{
    Push-T planning efficiency.
    (a) Success on medium-length tasks versus mean time per planning
    call as the CEM budget varies, for generic initialization and for
    priors fine-tuned on 10, 20, and 30 demonstrations; lines are
    saturating-exponential fits.
    (b) CEM objective over iterations with and without policy
    initialization on identical planning problems from 90 held-out
    initial states; costs are normalized to generic initialization at
    iteration one, and shading shows standard errors.
    }
    \label{fig:pusht_efficiency}
\end{figure}

\textbf{Planning efficiency.}
Fig.~\ref{fig:pusht_efficiency}(a) plots medium-task success against
mean time per planning call as the CEM budget (iterations and samples)
varies. The prediction horizon and subgoal schedule are fixed.
We compare generic initialization with priors fine-tuned on 10, 20,
and 30 demonstrations; the 30-demonstration prior is also used in
Table~\ref{tab:pusht_main}. All three priors improve the trade-off
between success rate and planning time. The 30-demonstration prior
exceeds 60\% success at 8\,s per call, whereas generic initialization
reaches 33\% at 50\,s. Even the 10-demonstration prior, whose policy
alone succeeds on 3.3\% of medium tasks, raises the fitted plateau from
34\% to 48\%. The policy query adds under 0.2\,s per call.
Panel~(b) compares the two initializations on identical planning
problems. Starting from similar first-iteration costs, policy
initialization reaches in four iterations the cost that generic
initialization attains in ten, and ends with a 24.1\% lower cost under
the same sampling budget.

\textbf{Generated versus oracle subgoals.}
PILOT (oracle) selects $K$ frames from the corresponding held-out expert
trajectory using the same normalized temporal sampling rule as for
generated videos. Both variants share $K$, per-subgoal execution budget
$R$, prediction horizon, policy prior, and CEM settings.
Expert subgoals improve success over generated subgoals by 10.0, 30.0,
and 30.0 percentage points on short, medium, and long tasks,
respectively. This gap suggests room to improve generated subgoals
under the same controller. Possible causes include videos that fail to
depict task completion, limited visual fidelity, unreachable subgoals,
and temporal misalignment.

\begin{figure}[t]
\centering
\includegraphics[width=\columnwidth]{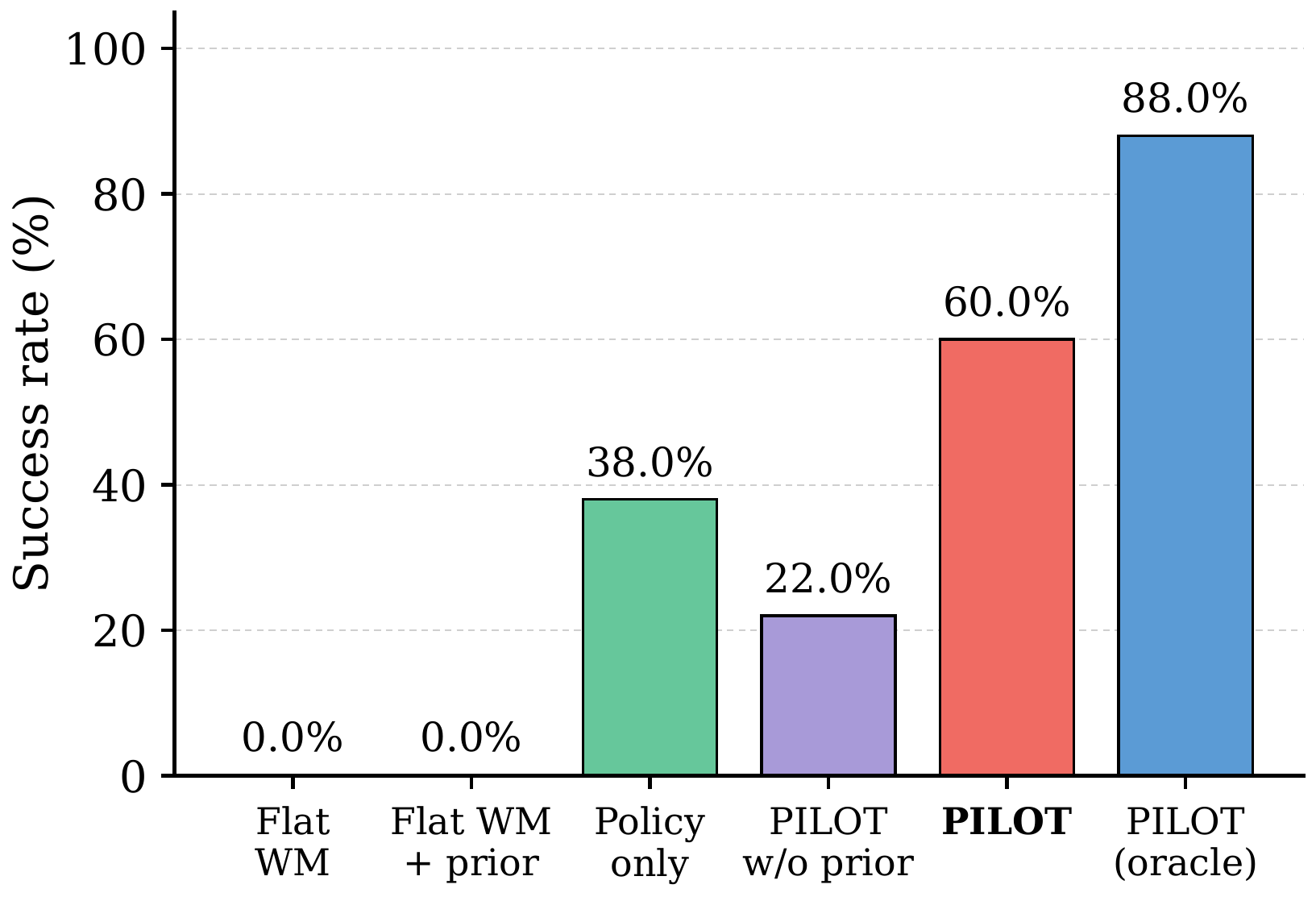}
\caption{
LIBERO-10 success rate
(\NLIBERO{} held-out initial states per task, aggregated over the ten
tasks). PILOT (oracle) uses expert subgoals.
}
\label{fig:libero_main}
\end{figure}

\textbf{Long-Horizon Manipulation.}
\begin{figure*}[!t]
    \centering
    \includegraphics[width=\textwidth]
    {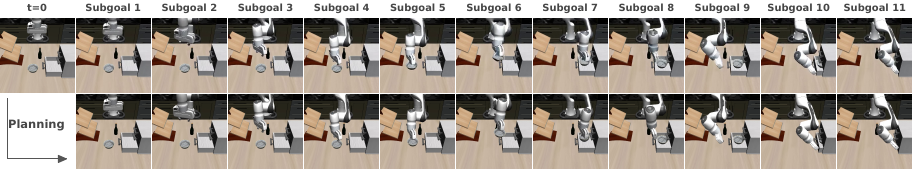}
    \caption{
    PILOT on the LIBERO-10 task
    \emph{put the black bowl in the bottom drawer of the cabinet and
    close it}.
    Top: initial observation and video-predicted visual subgoals.
    Bottom: simulator observations after the allotted control steps for
    each subgoal.
    }
    \label{fig:libero_subgoal_vs_exec}
\end{figure*}
We evaluate all six variants on \NLIBERO{} held-out initial states per
LIBERO-10 task. Fig.~\ref{fig:libero_main} reports success rates
aggregated over the ten tasks, and
Fig.~\ref{fig:libero_subgoal_vs_exec} shows a successful execution.

Flat WM achieves 0.0\% success with either initialization.
Generated subgoals with generic CEM initialization yield 22.0\%
success, below the 38.0\% achieved by direct policy execution.
Combining subgoals and the policy prior, PILOT reaches 60.0\%, a
22 percentage-point gain over Policy only. Expert subgoals raise success
to 88.0\%, suggesting that better temporal guidance could further
improve execution.

\ifarxiv
\subsection{World-Model Diagnostics}
\label{sec:wm_diagnostics}

We examine how JEPA-WM predictions and planning costs respond to
candidate controls. As shown in
Fig.~\ref{fig:wm_rollout_comparison}, the reference and perturbed action
sequences produce visibly different predicted trajectories from the
same initial observation, illustrating the dynamics model's action
sensitivity.

\begin{figure}[t]
    \centering
    \includegraphics[width=\columnwidth]
    {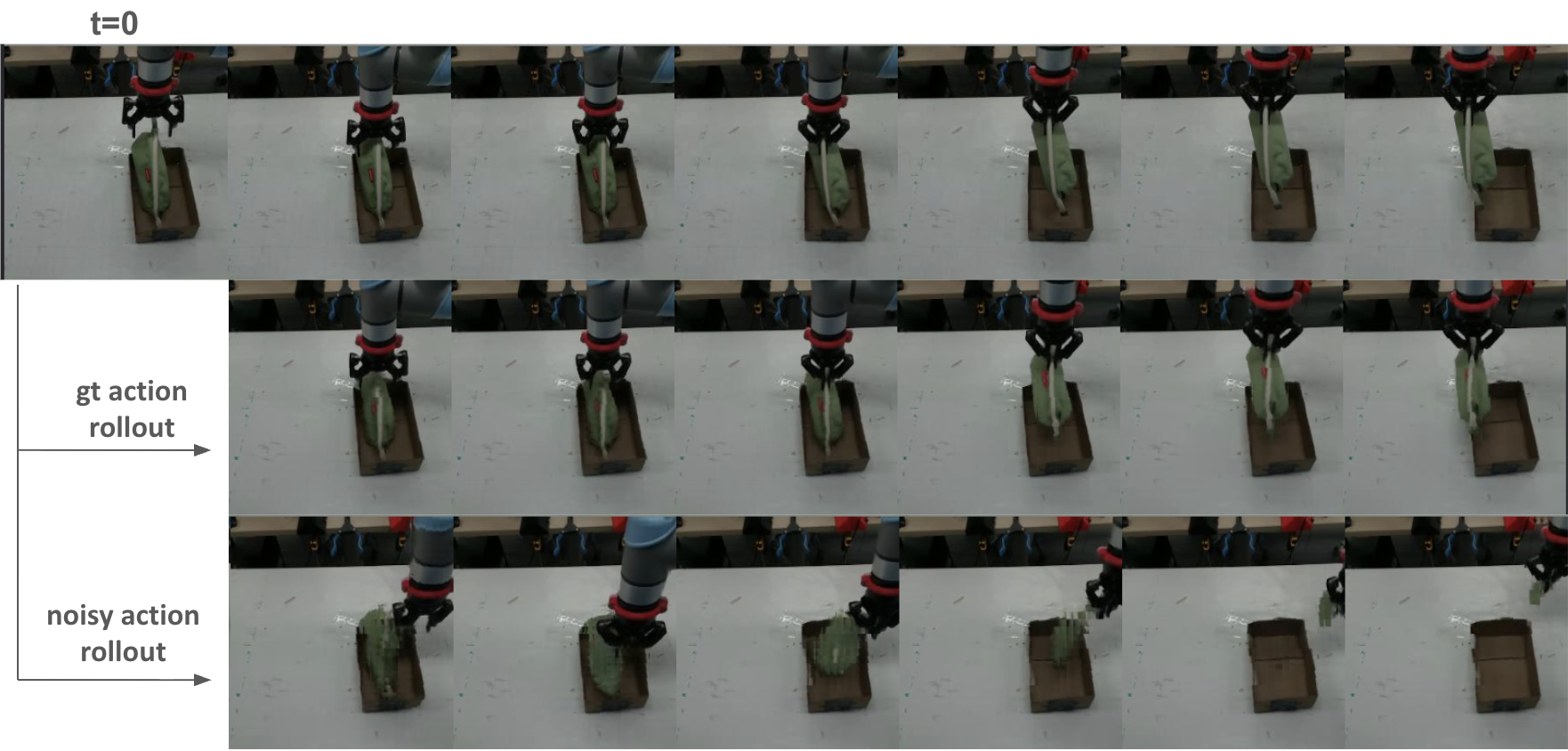}
    \caption{
    Action-conditioned world-model rollouts from the same initial
    observation.
    The reference action sequence produces task-relevant predicted
    motion, while a perturbed action sequence produces a different
    future trajectory, illustrating the action sensitivity of the
    learned dynamics.
    }
    \label{fig:wm_rollout_comparison}
\end{figure}

Fig.~\ref{fig:wm_energy_eval} shows two slices of the terminal latent
cost. The Cartesian slice fixes $\Delta x=0$ and sweeps
$(\Delta y,\Delta z)$; the gripper slice fixes Cartesian motion and
varies the gripper command. Both exhibit structured cost variation over
candidate actions.

\begin{figure*}[!t]
    \centering
    \begin{minipage}[t]{0.44\textwidth}
        \centering
        \includegraphics[width=\linewidth]{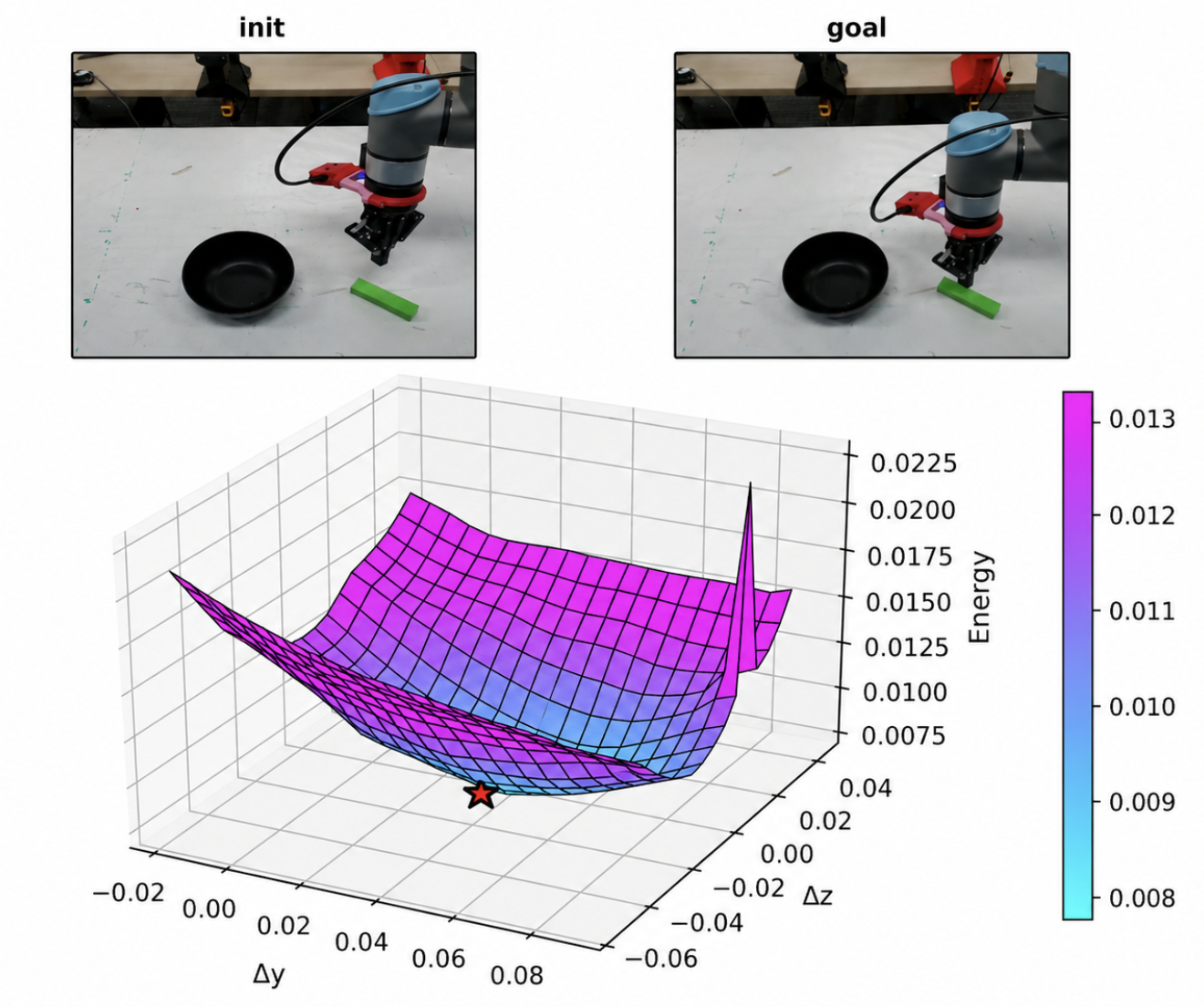}
        \small (a) Cartesian action slice with $\Delta x=0$.
    \end{minipage}
    \hfill
    \begin{minipage}[t]{0.44\textwidth}
        \centering
        \includegraphics[width=\linewidth]{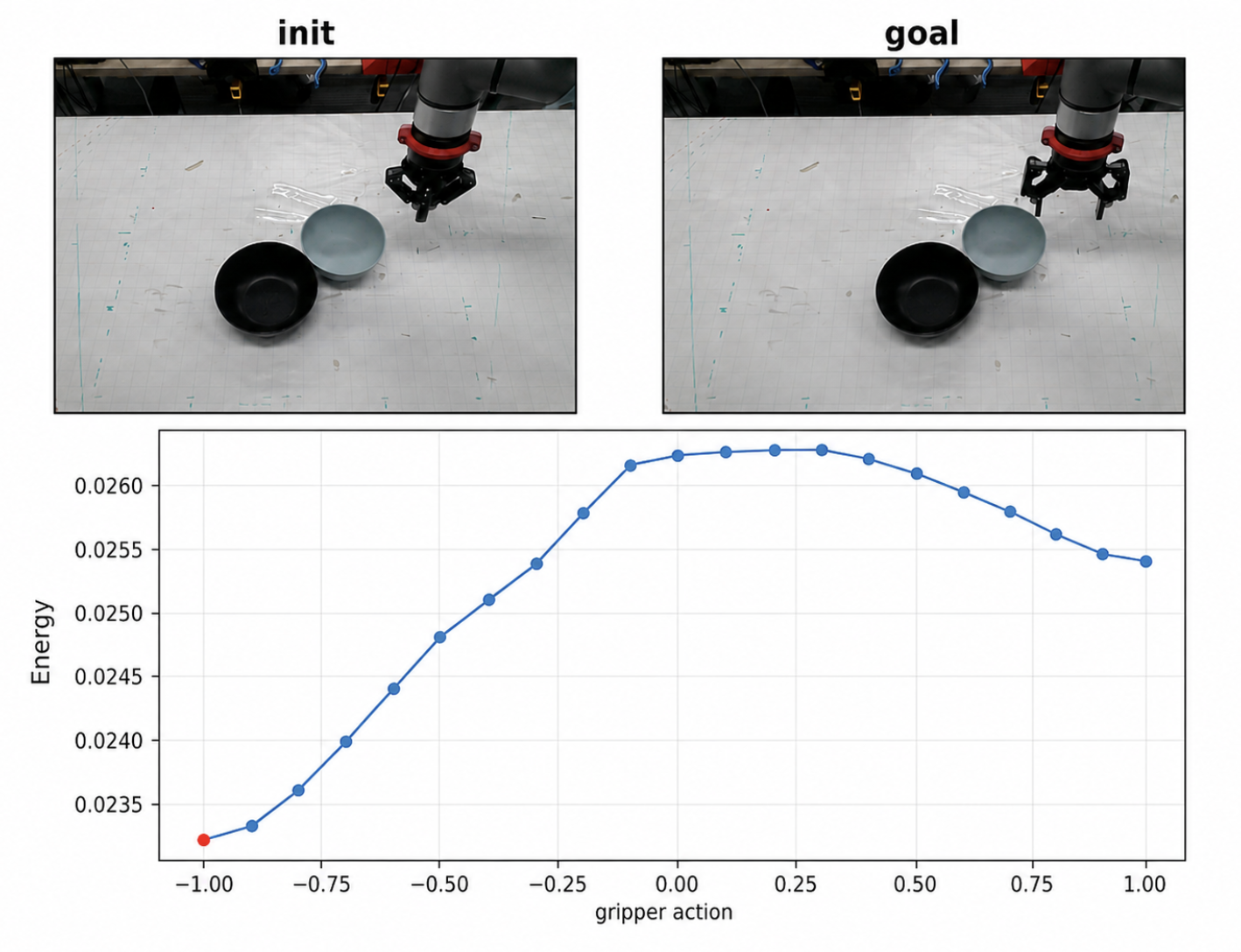}
        \small (b) Gripper action slice with fixed Cartesian motion.
    \end{minipage}
    \caption{
    World-model energy landscapes used for action evaluation.
    The top images show the initial observation and visual goal.
    The red marker denotes the selected low-energy action in each
    diagnostic slice.
    Lower energy corresponds to a predicted terminal latent state closer
    to the goal representation.
    }
    \label{fig:wm_energy_eval}
\end{figure*}
\fi

\section{CONCLUSIONS}

We presented \textbf{PILOT}, which combines video-predicted subgoals with policy-initialized local world-model planning. With Wan and $\pi_0$ fine-tuned on the same 40 real-robot demonstrations, PILOT achieves 68.8\% success across four tasks. The ablations show that visual subgoals can guide online control when direct policy execution is unreliable, while policy initialization improves local search. Simulation results further support these complementary roles: PILOT exceeds direct policy execution by 30 percentage points on long Push-T tasks and 22 points on LIBERO-10. Expert-subgoal comparisons suggest that better temporal guidance could yield further gains.

\section{Limitations and Future Work}

PILOT can fail when generated subgoals are unreachable, world-model errors misrank actions, or policy-centered sampling limits exploration. Its fixed subgoal schedule cannot accommodate delayed interactions, which can hinder recovery after a failed grasp. Video generation and online optimization also add latency relative to direct policy execution. Our real-world evaluation is limited to four tabletop tasks on a UR5e. Future work will investigate adaptive subgoal timing and search distributions, more efficient world-model rollouts, and generalization to more diverse tasks and embodiments.




\section*{ACKNOWLEDGMENT}

We thank Cloudrift for providing GPU compute credits and Srinivas Kantha Reddy for hardware support.

\bibliographystyle{IEEEtran}
\bibliography{example}

\end{document}